\documentclass[letterpaper, 10 pt, conference]{ieeeconf}  

\IEEEoverridecommandlockouts                              

\usepackage{graphics} 
\usepackage{epsfig} 
\usepackage{mathptmx} 
\usepackage{times} 
\usepackage{amsmath} 
\usepackage{amssymb}  

\title{\vspace{-1.5cm} 
{\footnotesize This work has been submitted to the IEEE for possible publication. Copyright may be transferred without notice, after which this version may no longer}\\
[-8pt]
{\footnotesize be accessible.}\\ 
\vspace{0.5cm} 
\LARGE \bf
Vision-Language Model-based Physical Reasoning for Robot Liquid Perception
}

\author{Wenqiang Lai\textsuperscript{1}, Yuan Gao\textsuperscript{1,}\textsuperscript{2}, Tin Lun Lam\textsuperscript{1,}\textsuperscript{2,}\textsuperscript{\dag}
\thanks{\textsuperscript{1}Shenzhen Institute of Artificial Intelligence and Robotics for Society.}
\thanks{\textsuperscript{2}School of Science and Engineering, The Chinese University of Hong
Kong, Shenzhen.}
\thanks{\textsuperscript{\dag}Corresponding author is Tin Lun Lam tllam@cuhk.edu.cn}
}

\begin{document}

\maketitle
\thispagestyle{empty}
\pagestyle{empty}

\begin{abstract}
There is a growing interest in applying large language models (LLMs) in robotic tasks, due to their remarkable reasoning ability and extensive knowledge learned from vast training corpora. Grounding LLMs in the physical world remains an open challenge as they can only process textual input. Recent advancements in large vision-language models (LVLMs) have enabled a more comprehensive understanding of the physical world by incorporating visual input, which provides richer contextual information than language alone. In this work, we proposed a novel paradigm that leveraged GPT-4V(ision), the state-of-the-art LVLM by OpenAI, to enable embodied agents to perceive liquid objects via image-based environmental feedback.  Specifically, we exploited the physical understanding of GPT-4V to interpret the visual representation (\textit{e.g., time-series plot}) of non-visual feedback (\textit{e.g., F/T sensor data}), indirectly enabling multimodal perception beyond vision and language using images as proxies. We evaluated our method using 10 common household liquids with containers of various geometry and material. Without any training or fine-tuning, we demonstrated that our method can enable the robot to indirectly perceive the physical response of liquids and estimate their viscosity. We also showed that by jointly reasoning over the visual and physical attributes learned through interactions, our method could recognize liquid objects in the absence of strong visual cues (\textit{e.g.}, container labels with legible text or symbols), increasing the accuracy from 69.0\%—achieved by the best-performing vision-only variant— to 86.0\%.
\end{abstract}

\section{Introduction}
How would a human respond to a query like ``\textit{Bring me the milk please}''? Intuitively, humans would perceive visually the environment to look for the queried object. In case the object cannot be distinguished from visual information alone, we perform additional observations and reasoning over information from other modalities.  For an intelligent robot to achieve human-like reasoning, the understanding of the feedback from the interactions in the physical world is essential~\cite{lake2017building}.
\begin{figure}[!t]
    \centering   \includegraphics[width=0.99\columnwidth]{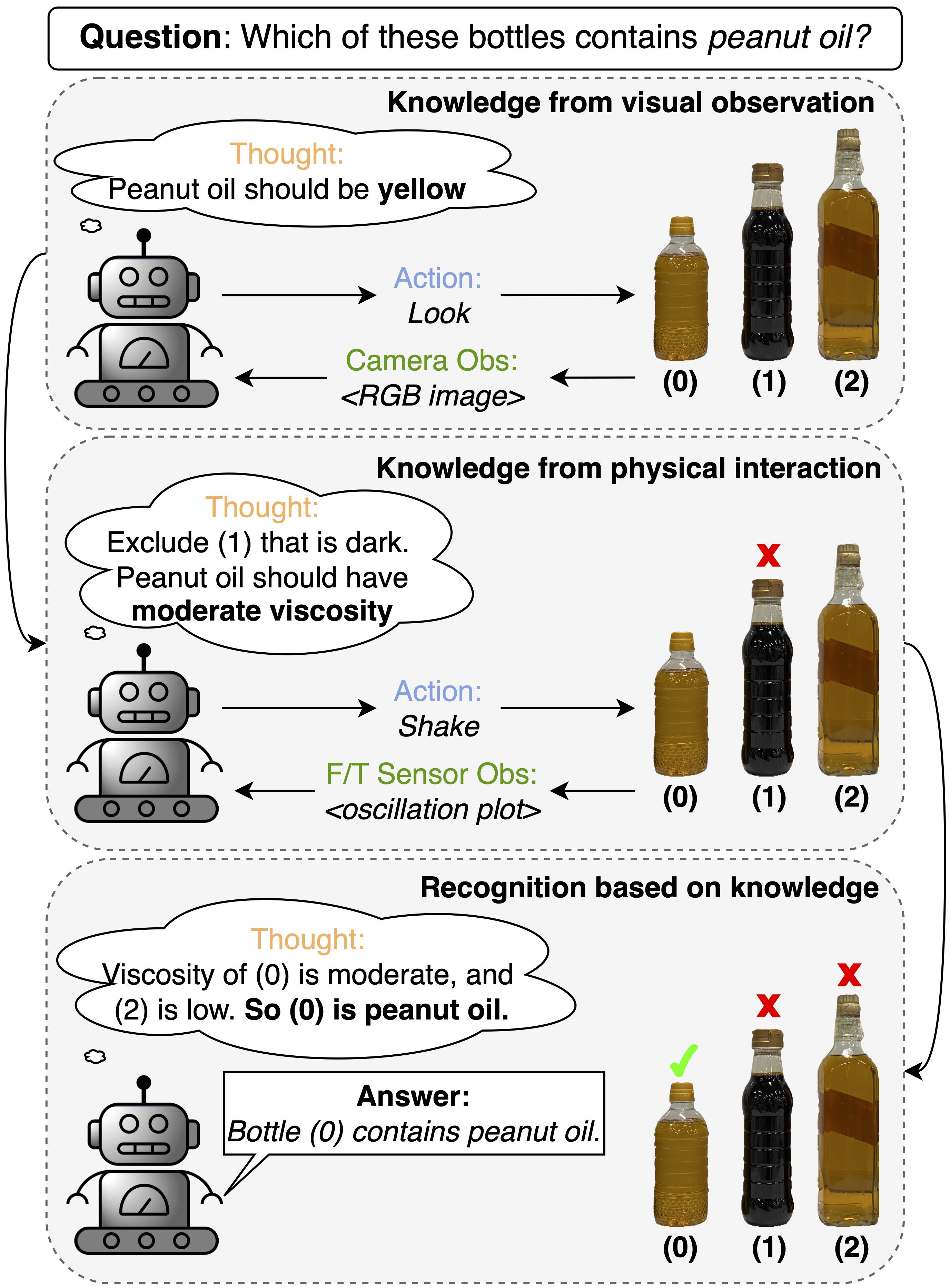}
    \caption{Our method firstly predicts the visual and physical properties of the queried liquid object using commonsense knowledge (bolded text), and then explores liquids via \textit{Look} and \textit{Shake} actions interactively to estimate their properties from image-based feedback. Finally, the liquid with most consistent properties is selected as the answer.}
    \label{fig:interaction_diagram}
\end{figure}
In recent years, large language models (LLMs) have shown remarkable performance on commonsense and physical reasoning tasks \cite{bisk2020piqa}, \cite{talmor2022commonsenseqa}, making them potential reasoning models for robotic tasks, such as planning and manipulation~\cite{huang2022inner}. To ground LLMs, prior works mainly relied on external modules to convert the multimodal environmental feedback into text. For example, ~\cite{zhao2023chat} exploited large-language models (LLMs) to recognize the material of objects in a simulated environment by reasoning over textualized multimodal interaction feedback. However, the lack of direct perception of the environment limited the comprehension of the world. More recently, large vision-language models (LVLMs), such as GPT-4 V(ision) by OpenAI~\cite{yang2023dawn}, enabled vision as an extra input modality in addition to language, which facilitated the incorporation of richer semantic knowledge. This opens up an interesting question: can LVLMs be used to reason over multimodal feedback from interactions in the physical world for robotic tasks? In this work, we used GPT-4V as our LVLM backbone model to explore a paradigm that adds image-based interaction feedback into reasoning the process of the LVLM to ground it within the physical world. Inspired by~\cite{yao2022react}, we allow the robot to act on a target object, and record the feedback in the form of images. Then, we feed the description of the taken action and the image-based feedback to the LVLM for inference. In this work, we evaluated this action-observation-reasoning paradigm with liquid objects, which require physical understanding to correctly recognize them, because the visual attributes (\textit{e.g.}, color and texture) of liquid objects are either sometimes unobservable due to the opacity of the containers, or insufficient to distinguish a liquid from visually similar ones. For example, to distinguish a target liquid object from visually similar counterparts, the robot would shake the bottles, and reason over the time-series plots of liquid oscillations using its physical understanding to provide a qualitative estimate of their viscosity and select the one with the most consistent properties (Fig.~\ref{fig:interaction_diagram}). 

We conducted a systematic evaluation of the proposed method on 10 common household liquids of varying container appearance, shape, and material. We established two settings with these liquids to simulate the variability in the appearance of liquids in the real-world. In the first setting, the original packaging labels remained intact, while these were removed or rendered invisible in the second setting.  We used a robot arm as the embodiment and a wrist-mounted F/T sensor and a RGB camera to provide feedback.
The main contributions of this work are summarized as follows:
\begin{itemize}
    \item We explored a new paradigm that leverages LVLM to perceive and reason over physical response from liquid objects via image-based haptic feedback for the qualitative estimation of liquid viscosity. 
    \item We demonstrated that by integrating both visual and haptic feedback, our method increased the accuracy to 86.0\% in recognizing 10 common household liquids, compared to the 69.0\% accuracy achieved by the variant using only visual feedback..
\end{itemize}

\section{Related Work}
\subsection{Large Models for Physical Reasoning}
The knowledge of the physical properties of an object is crucial in many robotic tasks. Prior works studied learning-based methods for the estimation of physical properties from visual \cite{wu2015galileo},\cite{fragkiadaki2015learning} and other modalities~\cite{wang2020swingbot},\cite{guo2023estimating} of interaction data. However, these methods are task-specific and difficult to scale, as they require a substantial amount of training data. Recently, there has been an increasing interest in leveraging the rich world knowledge encoded in LLMs for reasoning tasks. 

LLMs, which reverse-engineer the world through the massive quantity of training text, achieved remarkable performance on a range of physical reasoning tasks~\cite{liu2022mind}. However, since LLMs can only process texts, previous works relied on external modules~\cite{huang2022inner},~\cite{singh2023progprompt}, \cite{brohan2023can} to provide textual description of the feedback from the environment. For example, \cite{zhao2023chat} proposed a framework for robot planning in a simulated environment, where the robot agent plans to act in the environment to gather textualized object-centric physical properties from multimodal perception modules. In these works, the information provided by perception modules was conveyed via language, which may not provide the necessary context to comprehend the world, leading to inaccurate reasoning. 

More recently, LVLMs were used to directly reason over the visual feedback to estimate a range of object-centric physical concepts, such as mass and deformability~\cite{gao2023physically},\cite{li2023can}. These methods work by relating visually observable attributes, such as object material or category, to the intrinsic physical properties. Liquid objects have highly variable visual attributes, making the estimation of their intrinsic properties (\textit{e.g.}, viscosity) challenging. In this work, we exploit the physical reasoning ability of LVLM to perceive the viscosity of liquids by reasoning over image-based haptic interaction feedback. 

\subsection{Robot Liquid Perception}
The variability of liquids, combined with the way containers can alter their perceived shape and texture, posed great challenges to general object detection methods based on vision. As such, prior works leveraged non-vision interaction-based approaches to perceive liquids~\cite{chitta2011tactile,guler2014s,matl2019haptic,huang2022understanding}. These methods perceive the response of liquids to external motions (\textit{e.g.}, shake, grasp, tilt, etc.) using various sensors, such as accelerometer, F/T sensor, and tactile sensor, based on which data-driven or physical analysis-based models were used for the estimation of physical properties (\textit{e.g.}, viscosity) and/or classification of liquids. 

Although data-driven methods~\cite{chitta2011tactile},\cite{guler2014s} reported promising results in liquid classification tasks, they lack of scalability and generalization to new liquids. In contrast, \cite{matl2019haptic} proposed a method based on fluid dynamic analysis to estimate the volume, mass, and viscosity of liquids. However, this method requires exact knowledge about the container's geometry, which is impractical in many scenarios. Later, \cite{huang2022understanding} combined physical analysis with a data-driven model to estimate the viscosity of liquids. The method works by firstly analyzing the dynamic tactile signals recorded after perturbations, and then using the extracted information (\textit{e.g.}, rate of damping and oscillation frequency) to train a regression model. However, this method requires re-training to generalize to novel liquid containers. 

Despite prior works achieved remarkable results, they either suffer from the generalization and scalability issues, or difficulty in the real-world deployment. In contrast to previous methods, we propose to leverage the commonsense and physical knowledge 
of LVLM to recognize liquid objects without any training or fine-tuning by reasoning over visual and haptic feedback.

\section{Methodology}
\subsection{System Architecture}
\addtocounter{footnote}{-1}
\begin{figure*}[!t]
    \centering   \includegraphics[width=2\columnwidth]{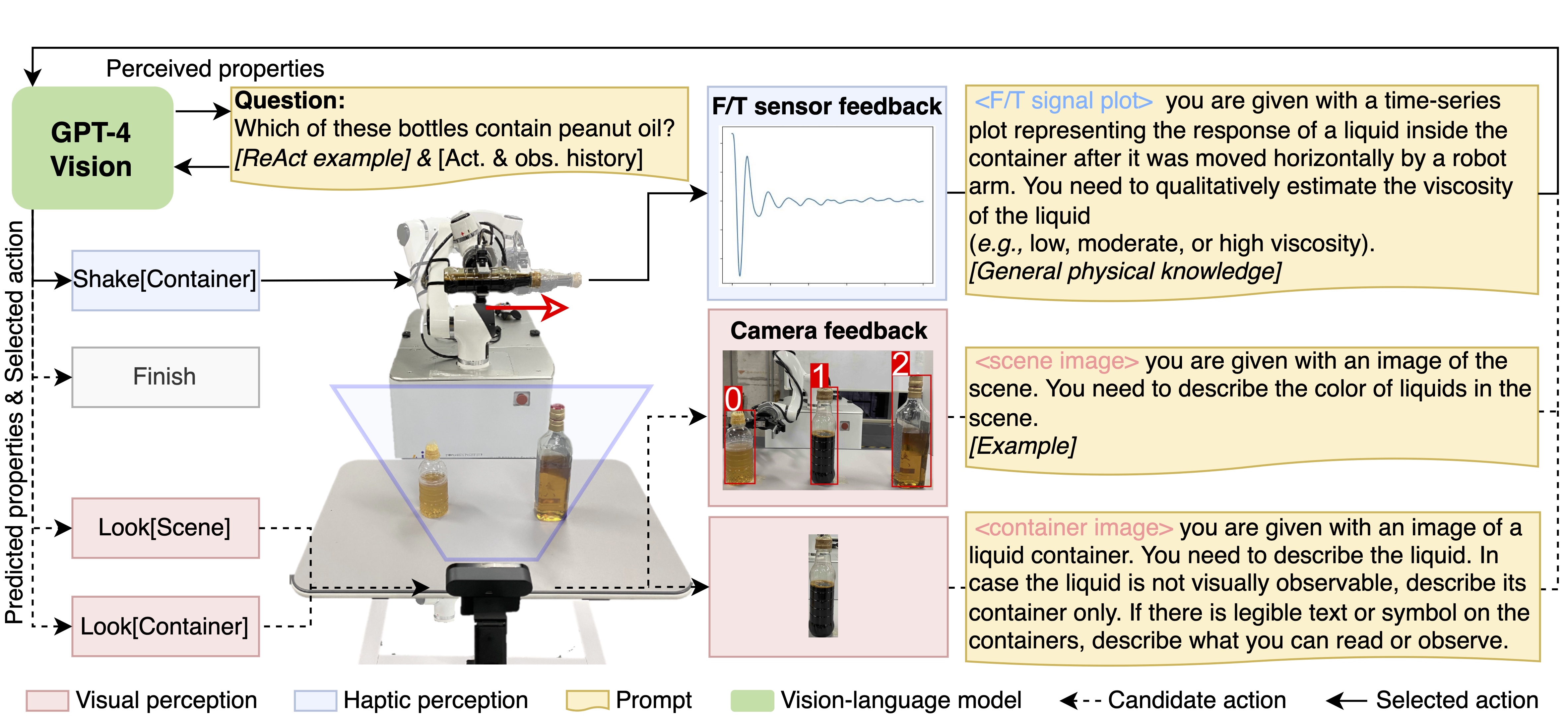}
    \caption{To recognize a target liquid object, we prompt (yellow) GPT-4V to reason, act and perceive in a closed-loop. It first predicts the properties of the target liquid object using prior knowledge, and selects an appropriate action to perceive the actual properties from haptic (blue) and visual (pink) feedback. Haptic feedback is denoised and converted into time-series plot, while visual feedback is pre-processed to include object bounding boxes. Based on the predicted and perceived properties, GPT-4V plans for next action to gather more information. GPT-4V terminates the loop with the action \textit{Finish} when the information is sufficient to recognize the target object. Dashed arrows represent potential action-reasoning paths not selected in the current loop.\protect\footnotemark}
    \label{fig:flowchart}
\end{figure*}

To enable interactive multimodal reasoning, we extended the reason-act (ReAct) framework proposed by~\cite{yao2022react} to allow the robot agent to act and perceive in the real world. Specifically for the liquid object perception and recognition tasks, we employed GPT-4V as an integrated perception, reasoning and planning model, and introduced an action space $A =$ \{\textit{Shake[Container], \textit{Look[Scene]}, \textit{Look[Container]}, \textit{Finish}}\}, where the first 3 actions are used to gather object-centric information, and the \textit{Finish} action is used to terminate the action-reasoning loop and return the final answer. The actions are described in detail in Section~\ref{sec:multimodal_perception}. 

Our method works by mimicking human behavior in recognizing a target object. Given a set of objects, humans use their \textit{internal world model} to predict the properties that the target object might possess, and then interactively explore the objects to gather multimodal feedback, from which the actual properties can be perceived. Then, the object with perceived properties that are consistent with the predicted properties is selected as the target object. Our method used LVLM as the \textit{internal world model} to recognize liquid objects, leveraging its commonsense and physical understanding.

Formally, let $obj$ denote the target liquid object, $q$ denote the prompting question (\textit{e.g.}, ``which of these bottles contains peanut oil?''), $e$ denote the in-context example to guide the reasoning-action behaviour~\cite{yao2022react}, and $c_t = (a_0, o_0, a_1, o_1, ..., a_{t-1}, o_{t-1})$ denote the context at time $t$, where $o$ represents the textualized perceived properties after executing action $a \in A$. We prompt LVLM to first predict the properties of the queried object $o'_t$ and then select an appropriate action $a_t$ to perceive the corresponding properties $o_t$, which can be considered as a mapping:
\begin{equation}
    F_{ReAct}:(q, e, c_t) \mapsto (o'_t, a_t)
    \label{eq:react}
\end{equation}
after executing $a_t$, we prompt LVLM to perceive the object properties by reasoning over the image-based feedback $I_t \in \mathcal{T}=$\{\textit{F/T signal plot}, \textit{scene image}, \textit{container image}\}, which can be regarded as visual question answering:
\begin{equation}
    F_{perception}:(q_I, I_t) \mapsto o_t
    \label{eq:perception}
\end{equation}
where $q_I$ is specific to the type of $I_t$ (Fig.~\ref{fig:flowchart}). The action-perception pair is then appended to the context in (\ref{eq:react}), $c_{t+1} = \{c_t, a_t, o_t\}$, to support the reasoning and acting in the next loop. A termination action \textit{Finish} is available for the robot to finally return the answer when the perceived properties of a particular liquid object match the predicted properties of the queried liquid object $o_t \sim o'_t, \forall t$.
\subsection{Multimodal Interactive Perception}
\label{sec:multimodal_perception}
To recognize liquid objects, complementing vision with haptic feedback from interaction is an intuitive approach for humans. Similarly, our method first observes the liquid objects using vision, and actively gathers additional physical knowledge about the objects via haptics.
\footnotetext{Codes and prompts at https://anonymous.4open.science/r/vlm\_reasoning}
\subsubsection{Vision}  Existing LVLMs, such as GPT-4V, tend to overlook fine details in high-resolution images and are prone to hallucinations when the scene is cluttered~\cite{yang2023dawn}. Therefore, we introduced actions \textit{Look[Scene]} and \textit{Look[Container]} to perceive coarse and fine-grained visual attributes, respectively. Color and shape are the most distinctive properties for object recognition~\cite{buetti2019predicting}, and since liquids are shapeless, we consider color as the coarse visual property to be observed through the action \textit{Look[Scene]}. We pre-processed the scene images using the pre-trained open object detector Owl-ViT~\cite{minderer2022simple} to provide visual reference (\textit{e.g.}, bounding boxes with index numbers on top) to the liquid objects, which has been shown to improve the visual question answering ability of LVLM~\cite{yang2023dawn}. To avoid hallucinations, we introduced the action \textit{Look[Container]} that allows the robot to select the interested object in the scene and observe it more closely. For simplicity, we implemented \textit{Look[Container]} as a command that crops the interested object from the scene image, following~\cite{wu2023textit}. Specifically, when the robot selects the action \textit{Look[Container]}, a cropped image of the target container will be returned, from which the robot can observe fine-grained details (\textit{e.g.}, transparency and legible text/symbols on the container) in the absence of distractors. To guide LVLM to generate the desired output format for \textit{Look[Scene]}, we provided a hypothetical example, in which a generic label (\textit{e.g.}, [Input Image]) replaces an actual image, to avoid information leakage during the evaluation. For the action \textit{Look[Container]}, we prompt LVLM to simply provide a comprehensive description of the object in the cropped image without any in-context example.

\subsubsection{Haptics} 
We used haptic feedback to capture the physical attributes, such as viscosity, of the liquid objects. Since our approach only focuses on reasoning over the haptic feedback from a high level, it is less sensitive to noises. Therefore, we employed a 6-DoF F/T sensor, rather than an expensive high-resolution tactile sensor used in~\cite{huang2022understanding}, to collect haptic feedback during the interactions with the liquids. We designed the action \textit{Shake[Container]} for the robot to shake the target container, and record the response of the liquid to the motion. 

Following \cite{huang2022language}, the robot uses its arm to move the target container, which is placed horizontally in the gripper, 10 cm along the axis connecting the bottom and the opening of the container. We employed a scripted motion since object manipulation is not the focus of this work. Since the motion was linear, and the force measurements were not sensitive enough to capture the motion patterns due to hardware limitations, we only recorded the torque measurements along the axis that is orthogonal to both the direction of motion and the gravitational force for 10 seconds while holding still the container. We applied a 5th-order low-pass filter with a cut-off frequency of 2 Hz to denoise the collected signal, and then normalized it to zero-mean and unit-variance following~\cite{guler2014s}. With the given context (\textit{e.g.}, the action taken), the LVLM should interpret the image feedback (\textit{e.g.}, plot of F/T sensor signal) using relevant physical understanding and provide an answer to the question. Humans require reference in reasoning about liquid viscosity. Likewise, we injected general physical knowledge, consisting of two descriptions of the expected oscillation patterns for low (\textit{e.g.}, peaks with slowly decreasing amplitudes) and high viscosity (\textit{e.g.}, peaks with rapidly decaying amplitudes) liquids to the prompt as references, following the experimental design in~\cite{van2018visual}, in which the maxima and minima stimuli were presented to the observers before asking them to rate the viscosity of liquids.

\section{Experiments}
We used GPT-4V by OpenAI\cite{yang2023dawn} as the LVLM backbone in our method. For more deterministic inference, we set the temperature parameter to 0. We conducted two experiments to answer the following questions: 1) Can GPT-4V correctly interpret the image-based haptic feedback using relevant physical understanding to estimate viscosity? 2) How does the available visual and haptic object-information affect the liquid recognition performance of GPT-4V? Our method was designed to work in an interactive manner in the physical world, however, we employed offline evaluation to save time for collecting haptic feedback in real-time to avoid system failures caused by irrelevant modules (\textit{e.g.},  grasp failures).
\subsection{Hardware} We used a single-arm robot with a mobile base by Moying Technology and a wrist-mounted 6-DoF F/T sensor by Robitq with a sampling frequency of 100 Hz. A parallel gripper by Robotiq is mounted on the wrist as the end-effector. A Logitech Brio camera was placed on a tripod at a suitable pose, such that all liquid objects could be captured within the image. The liquids evaluated in this work were \textit{coke, water, olive oil, peanut oil, soy sauce, whiskey, balsamic vinegar, orange juice, honey, milk}, which are commonly found in domestic environments with viscosity spanning from low (\textit{e.g.}, \textit{water}) to high (\textit{e.g.}, \textit{honey}). We measured the ground truth viscosity of the liquids using the NDJ-1 rotational viscometer. In contrast to most prior works that assumed containers of various liquids have the same shape and material, liquids evaluated in our experiments were contained in their original containers made of various materials (\textit{e.g.}, plastic, glass, and paper) with  various shapes and length. Relaxing this assumption is the key to the deployment in the real world, as in most cases the liquids have containers of variable geometry and appearance.
 \begin{figure}[!t]
    \centering
    \includegraphics[width=0.9\columnwidth]{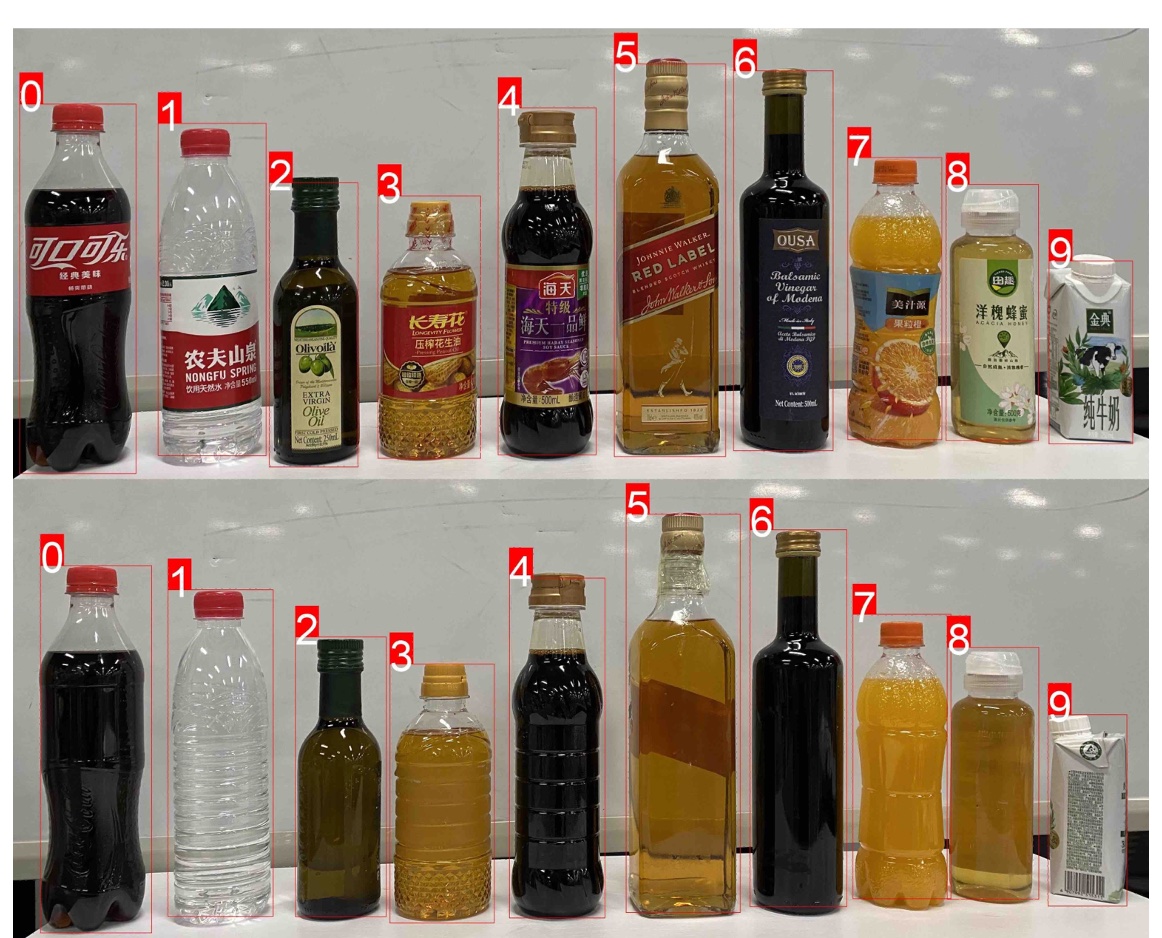}
    \caption{Ten common household liquids evaluated in the experiments placed in a line on a table, each surrounded by a bounding boxed indexed between 0 and 9 from left to right: \textit{coke, water, olive oil, peanut oil, soy sauce, whiskey, balsamic vinegar, orange juice, honey, milk}. (Top) Liquids in their original packaging with text and symbols on the labels. (Bottom) Same liquids with labels being removed or rendered invisible.}
    \label{fig:exp_liquids}
\end{figure}
\subsection{Data Collection} 
\begin{figure*}[!t]
    \centering   \includegraphics[width=1.99\columnwidth]{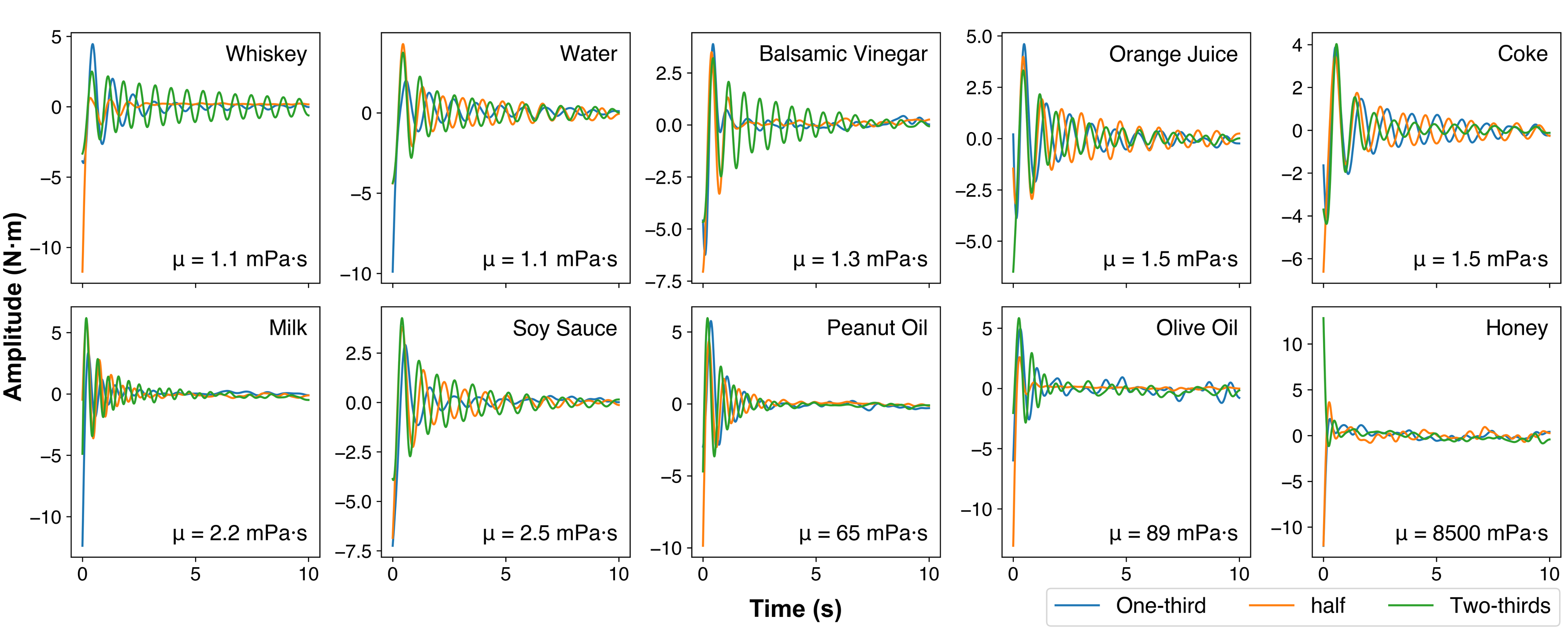}
    \caption{The plots of 10-second F/T sensor signals collected after shaking the containers of 10 common household liquids filled at \textit{one third}, \textit{half}, and \textit{two thirds} of capacity of their containers, sorted by their viscosity. Signals were filtered using a 5th order low-pass with a cut-off of 2 Hz and standardized.}
    \label{fig:all_plots}
\end{figure*}
We collected images of the scene with different settings and the haptic feedback of each liquid separately, and provided these directly to GPT-4V when the corresponding action was chosen. As shown in the image of the scenes (Fig.~\ref{fig:exp_liquids}), we established two distinct experimental settings to simulate the real-world variability of visual attributes in liquids and their containers, where one lacks control over the presentation of these objects. Since the haptic signal might be too weak for perception when the fill level of liquids is either too low or too high. Following~\cite{huang2022understanding}, we collected one haptic feedback from each liquid object at 3 fill levels in the middle range, namely \textit{one-third, half, and two-thirds}. Fig.~\ref{fig:all_plots} showed the plots of the collected haptic feedback data of each liquid at different fill levels, sorted in ascending order by their ground truth viscosity. The fill levels were measured against the capacity of the containers using a beaker. As we do not deal with object grasp pose and force estimation, we simply hard-coded a grasping pose and force for each liquid. Following \cite{guler2014s}, the scripted grasp poses were set to around the middle region of the containers. The pose and force were determined empirically to avoid slippage.

\begin{table}[!b]
\caption{Accuracy of GPT-4V in predicting pair-wise viscosity relation of liquids at three fill levels with plain and knowledge-enhanced prompts.}
\label{tab:viscosity}
\centering
\begin{tabular}{lcc}
\hline
& \multicolumn{2}{c}{\textbf{Accuracy (\%)}}\\ \cline{2-3}
\textbf{Fill level} &  Plain & Knowledge-Enhanced\\ \hline
One third     & 66.4 & 77.1  \\
Half          & 67.8 & 77.5 \\
Two thirds    & 66.4  & 79.9 \\ 
\hline
\end{tabular}
\vspace{5pt} 
\end{table}

\subsection{Relative Viscosity Estimation via Haptics}
To answer the first question, we tested the contextual haptic feedback understanding of GPT-4V by prompting it to observe the haptic feedback of two liquids and estimate their viscosity relation. Slight modifications were made to the haptic perception prompt in Fig.~\ref{fig:flowchart}, where the input image consisted of two plots concatenated horizontally, and a question (\textit{e.g.},\textit{``which one is more viscous?}) was appended at the end. To investigate how the injected physical knowledge affects the estimation of viscosity, we created and evaluated a prompt without external knowledge, which we refer to as plain prompt. We conducted 10 trials for each pair of liquids at each fill level. The accuracy was reported in Table~\ref{tab:viscosity}. Note that we excluded invalid outputs (\textit{e.g.}, where GPT-4V refused to answer) from the calculation of accuracy.  

As expected, we observed that the injection of knowledge increased the accuracy compared to the plain prompt across all fill levels. GPT-4V achieved the highest accuracy of 79.9\% with knowledge-enhanced prompt in estimating viscosity relation of two thirds-filled liquids, surpassing the result obtained using plain prompt by 13.5\%. There was not significant difference between accuracy across different fill levels with plain prompt, while there was a direct proportion between the accuracy and the fill level when using knowledge-enhanced prompt. We argue that this relation can be attributed to the differences in the shapes of containers. When the bottles are two thirds-filled, the impact of the shape of bottles would be minimal as the height of liquid when placed horizontally would be higher than the bottleneck. When the bottles are filled to one-third or half, the liquid’s sloshing motion will be affected by the shape of the bottleneck, which has a variable design across each liquid. 
\begin{figure}[!b]
    \centering   \includegraphics[width=0.99\columnwidth]{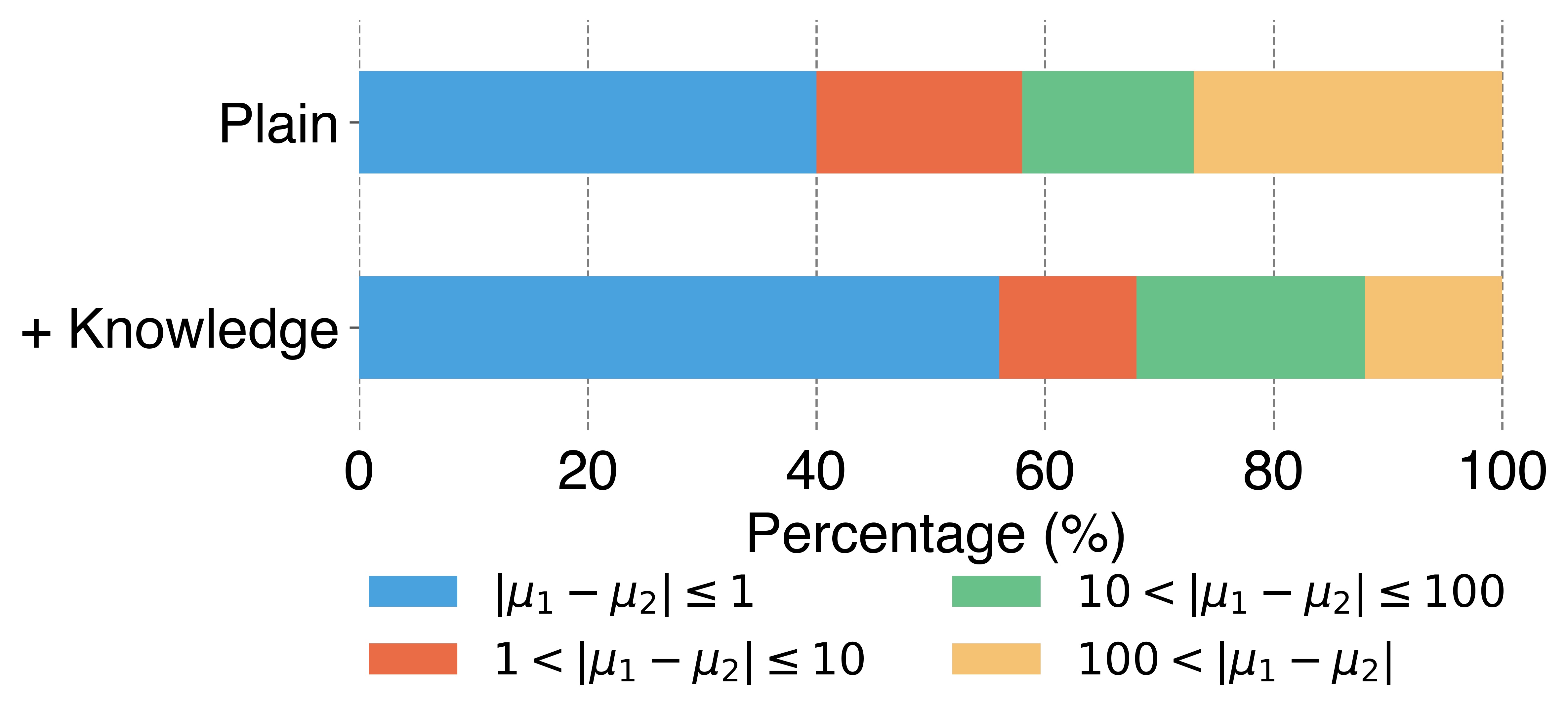}
    \caption{The error breakdown of plain and knowledge-enhanced prompts the absolute ground truth viscosity difference of liquid pairs.}
    \label{fig:error_breakdown}
\end{figure}
From the error breakdown (Fig.~\ref{fig:error_breakdown}), we observed that the accuracy increment of the knowledge-enhanced prompt was mainly driven by the increased performance in estimating pairs of liquids with large differences in ground truth viscosity. The percentage error caused by liquid pairs with an absolute viscosity difference larger than 100 mPas was largely reduced.

\subsection{Liquid Recognition via Multimodal Feedback}

\begin{table}[htbp]
\caption{Recognition accuracy comparison between our method and its variants with partial action space.}
\label{tab:vis_hap_exp}
\centering
\begin{tabular}{lc|cc}
\hline
& \multicolumn{2}{c}{\textbf{Accuracy (\%)}}\\ \cline{2-3}
\textbf{Methods / Settings} & \textbf{W/o labels} & \textbf{W/ labels}  \\ \hline
Look[Scn.]     & 62.0   & 76.0     \\
Look[Scn.]+Shake[Cnt.]  & 56.0 & 67.0   \\
Look[Scn.]+Look[Cnt.]        & 69.0      & \textbf{97.0}   \\ 
Look[Scn.]+Look[Cnt.]+Shake[Cnt.](Ours)    & \textbf{86.0}     & 93.0    \\
\hline
\end{tabular}
\vspace{5pt} 
\end{table}
We evaluated the performance of our method against its variants with partial action space (\textit{e.g.}, some exploration action is unavailable) to understand the impact of object-centric information gathered from different exploration actions. In the liquid recognition experiments, we have two settings with different visually perceivable cues. As shown in Fig.~\ref{fig:exp_liquids}, the first setting had the labels of all the containers facing the camera, so that the text and symbols on the labels can provide additional hints on the class of liquid. In the second setting (bottom), the labels of all the containers were removed to make sure the available visual cues were restricted to the container's color, shape, and material, as well as the color and texture of the content, in case of non-opaque containers. 
All liquids in the experiments were filled at around two-thirds of their respective capacity. We conducted 10 trials across all liquids in each setting. 

As shown in Table~\ref{tab:vis_hap_exp}, 
our method achieved a cognition accuracy of 86.0\% in the setting without labels on the containers, in which visual attributes were insufficient for correct recognition, outperforming all variants by large margins (17\% to 30\%), demonstrating the importance of physical knowledge. However, in both settings, we observed that the haptic perception combined with coarse visual perception (\textit{Look[Scene]} with \textit{Shake[Container]}) resulted in the lowest performance, even worse than using coarse visual perception alone. This might appear counter-intuitive at the first glance, however, since our method relied on coarse visual perception to select potential candidates for further interaction, misleading or insufficient information gathered in this first stage would strongly bias the future reasoning traces. This is consistent with the fact that the inclusion of fine-grained visual attributes largely increased the performance of all methods with action \textit{Look[Container]}. 

\begin{figure}[!t]
    \centering   \includegraphics[width=0.45\textwidth]{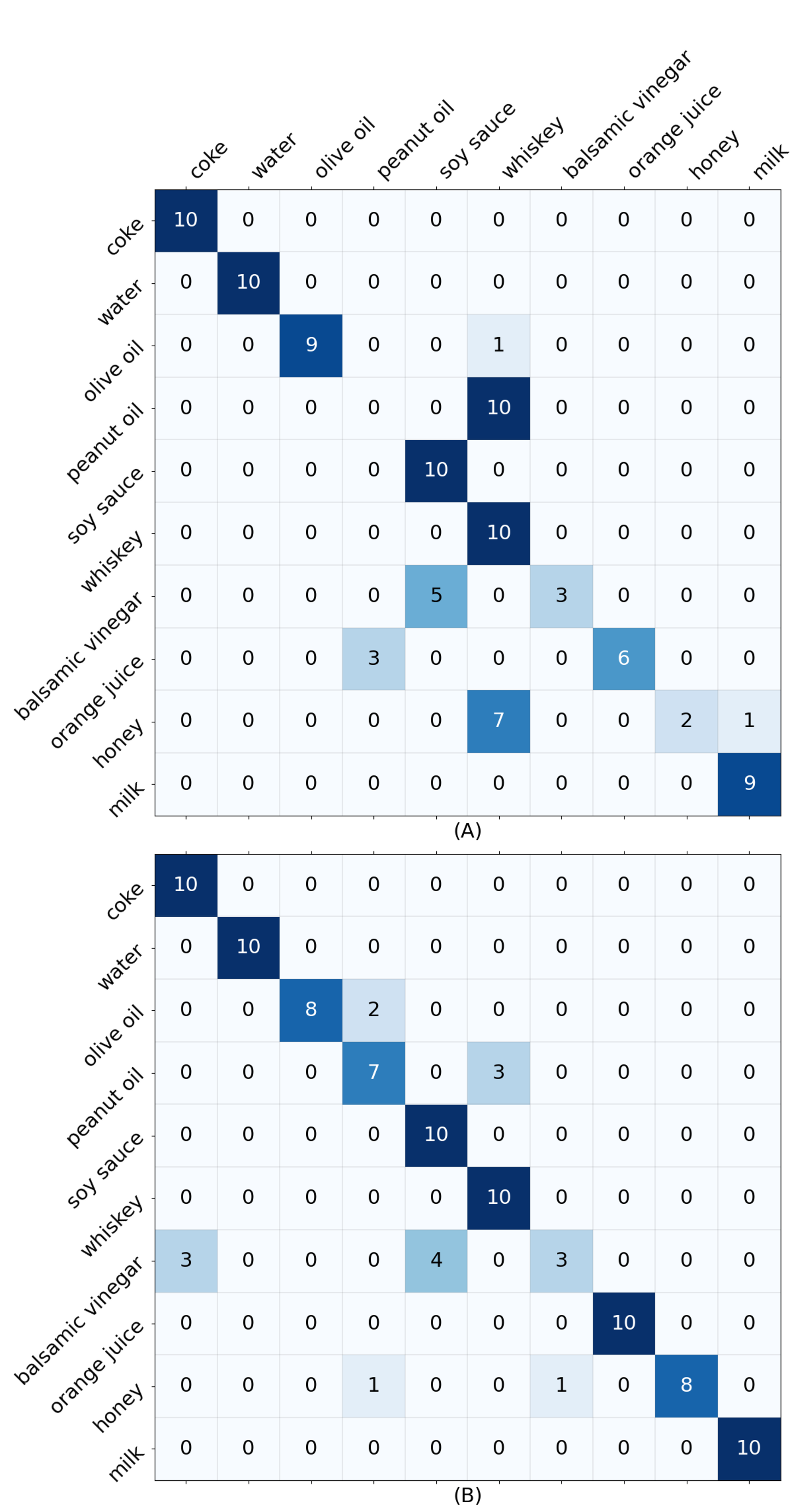}
    \caption{The confusion matrices for the recognition results of 10 liquids from the method with only two visual perceptual actions (A) and our methods (B) in the setting without labels. 10 trials were conducted for each liquid, and invalid results (\textit{e.g.}, where GPT-4V failed to give a definite answer) were excluded.}
    \label{fig:cm}
\end{figure}

Interestingly, in the setting with labels on the containers, we observed a small accuracy reduction of 4.0\% in our method when compared to the vision-only variant with two visual perceptual actions, which we investigated in a case study later.  The detailed breakdown of the predictions from our method and its variant without haptic feedback in the setting without labels on the containers was given in Fig.~\ref{fig:cm}. The major limitation of the vision-only method was the confusion within liquids with similar colors. For example, \textit{peanut oil} was confused with \textit{whiskey}, and \textit{honey}, while \textit{soy sauce} was confused with \textit{balsamic vinegar}. Whereas our method successfully allowed correct recognition of visually similar liquids with distinct viscosity (\textit{e.g.}, \textit{whiskey}, \textit{peanut oil}, \textit{honey}). However, no improvement was observed in the recognition of liquids with similar appearance and viscosity (\textit{e.g.}, \textit{soy sauce} and \textit{balsamic vinegar}).

\subsection{Case Studies}
\begin{figure*}[htbp]
    \centering   \includegraphics[width=1.99\columnwidth]{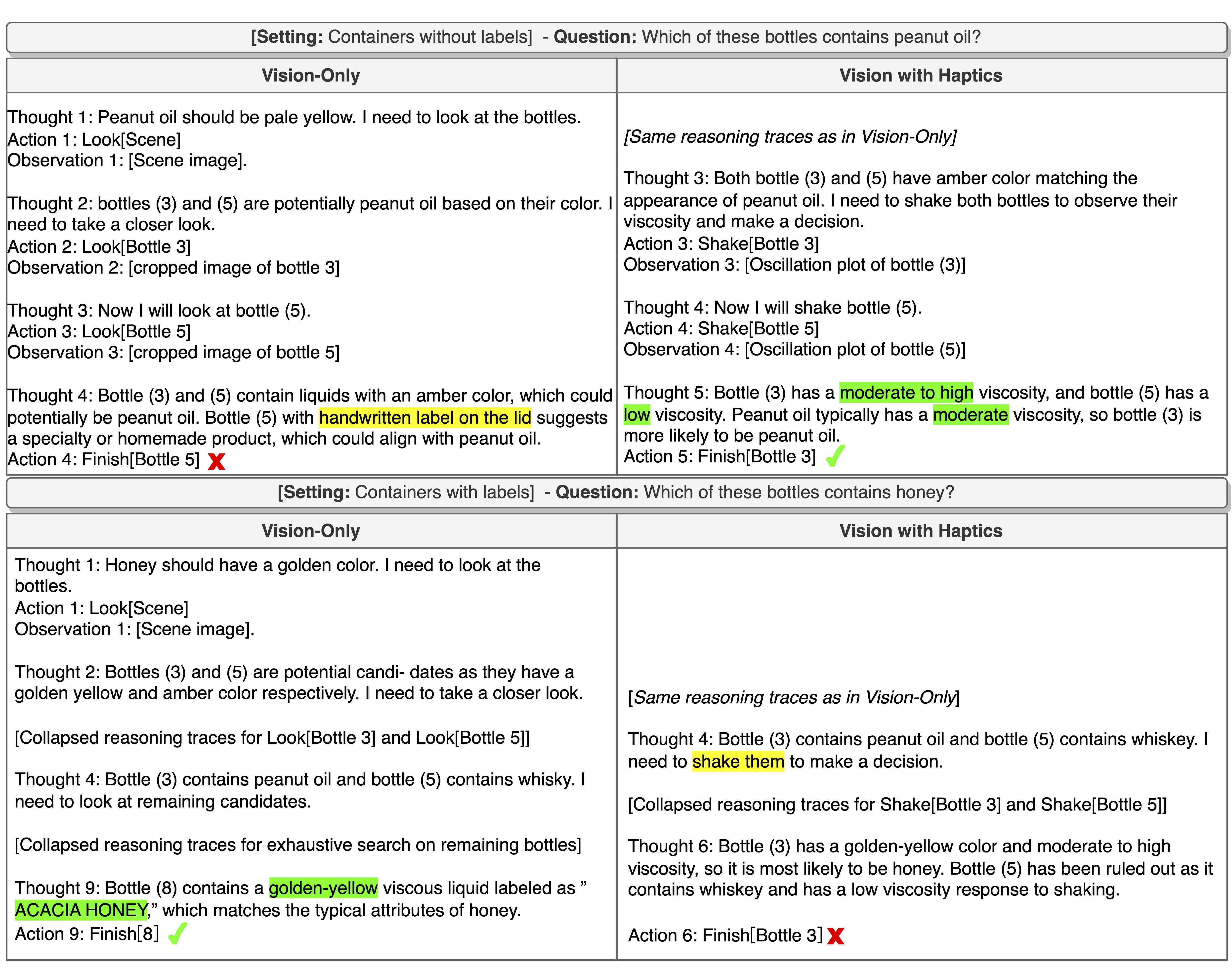}
    \caption{Reasoning traces of Vision Only method and Vision with Haptics method in recognizing (top row) peanut oil in the setting without labels on the containers, and honey in the setting with labels on the containers (bottom row). Misleading information were marked yellow, while correct information were marked green.}
    \label{fig:case_studies}
\end{figure*}
To understand how haptic feedback perception and physical understanding influenced the recognition of liquids across different settings, we provided the two case studies. Fig.~\ref{fig:case_studies} (top) showcased how the incorporation of haptic feedback combined with physical understanding contributed to the separation of visually similar liquid objects in the setting without labels. In the reasoning process to recognize \textit{peanut oil}, the vision-only method successfully identified bottle (3), which was the correct answer, as a potential candidate along with bottle (5), which contains \textit{whiskey}, because these have an amber color that is typical for \textit{peanut oil}. The vision-only system observed the handwritten label on the lid of bottle (5), which was assumed to be evidence of a homemade product. Without additional information, bottle (5) was wrongly returned as the answer. In contrast, with the incorporation of haptic feedback, the robot correctly recognized bottle (3) as \textit{peanut oil}, which has a moderate to high viscosity that is more consistent with the knowledge of GPT-4V about \textit{peanut oil} compared to bottle (5) that exhibited low viscosity. 

As seen previously, the inclusion of haptic feedback caused decrements in accuracy in the setting where there are clear symbols and legible texts on the containers. 
Fig.~\ref{fig:case_studies} (bottom) showcased the reasoning traces in recognizing \textit{honey}, which mainly contributed to the accuracy decrement of our method in the setting with labels. We observed that both methods wrongly identified bottle (3) and (5) as potential candidates from the initial observation of the scene, failing to include the correct answer. Upon closer observations, bottle (3) and (5) were discovered to contain \textit{peanut oil} and \textit{whiskey}, respectively. The vision-only method then proceeded with an exhaustive visual search, and finally found bottle (8) was labelled "Acaia Honey" and had a golden-yellow color that matched typical honey. Whereas our method proceeded to \textit{Shake} the candidates, ignoring the observed inconsistencies, and finally recognized bottle (3) as \textit{honey}, because of its moderate to high viscosity response. From this failed trial, we observed that our method is limited at distinguishing both visually and physically similar liquid objects, such as \textit{peanut oil} and \textit{honey}. The reason was that GPT-4V can only estimate the viscosity of liquids qualitatively, which was insufficient to describe the difference between \textit{peanut oil} and \textit{honey}. 

\section{Discussion}
The state-of-the-art LVLM, such as GPT-4V employed in this work, could potentially serve as the reasoning model for more generalized intelligent robot systems. As the reasoning traces revealed, GPT-4V can effectively reason over the image-based response of a liquid using its physical understanding to roughly estimate the viscosity, similar to humans whose perceptional sensing ability only allows qualitative description of the physical properties. Although the viscosity estimation is only qualitative, it is sufficient for humans to separate liquids that are visually similar but differ largely in viscosity in our daily life. In line with our expectations, GPT-4V exhibited similar reasoning traces in the liquid perception and recognition tasks as seen above. 
\section{Conclusion and future works}
LVLM possesses a strong visual and language understanding ability. The added vision modality facilitated the perception of the physical world, which in turn enhanced the reasoning ability of LVLM. In this work, we demonstrated how LVLM can be used to reason about the physical properties of liquid objects, and recognize them in an interactive manner. We proposed a method that allows LVLM to act in the physical world to gather multimodal feedback, and evaluated it on a liquid recognition task, which is a challenging problem in robotics. The experimental results suggested that GPT-4V was capable of combining its physical understanding to qualitatively estimate the viscosity of liquids from their image-based physical responses. Further, leveraging the knowledge about common liquid objects gained from pre-training, GPT-4V could recognize them by reasoning over the perceived visual and physical properties. The major limitation of this work is that GPT-4V is subject to continual updates that can affect model behavior, introducing a degree of unpredictability in the performance. Future works should consider evaluating a broader range of LVLMs to validate our approach's efficacy. While this work only focused on the viscosity estimation of liquid objects using only F/T sensor feedback, we highlight the potential of extending our method to more modalities for the reasoning of intrinsic properties of objects. For example, observing the visual representations (\textit{e.g.}, spectrograms) of auditory interaction feedback (\textit{e.g.}, sound reflection of an object), the material composition may be uncovered, facilitating downstream robotic tasks.


\addtolength{\textheight}{-12cm}   



\bibliographystyle{IEEEtran}
\bibliography{references}

\end{document}